\documentclass[10pt,twocolumn,letterpaper]{article}

\usepackage{cvpr}
\usepackage{times}
\usepackage{epsfig}
\usepackage{graphicx}
\usepackage{amsmath}
\usepackage{amssymb}

\usepackage{subfigure}
\usepackage{multirow}
\usepackage{multicol}
\usepackage{amsmath, bm}
\usepackage[linesnumbered, ruled]{algorithm2e}

\usepackage[breaklinks=true,bookmarks=false]{hyperref}

\cvprfinalcopy 

\def\cvprPaperID{****} 

\begin{document}

\title{Self-Distribution Binary Neural Networks}

\author{Ping Xue$^1$, Yang Lu$^{1,2}$, Jingfei Chang$^1$, Xing Wei$^{1,2}$, Zhen Wei$^{1,2}$\\\\
	
$^1$School of Computer Science and Information Engineering,\\Hefei University of Technology, Hefei 230009, China\\
$^2$Engineering Research Center of Safety Critical Industrial Measurement and Control Technology,\\ Ministry of Education Hefei University of Technology, Hefei 230009, China\\

}

\maketitle

\begin{abstract}
   In this work, we study the binary neural networks (BNNs) of which both the weights and activations are binary (\ie, 1-bit representation). Feature representation is critical for deep neural networks, while in BNNs, the features only differ in signs. Prior work introduces scaling factors into binary weights and activations to reduce the quantization error and effectively improves the classification accuracy of BNNs. However, the scaling factors not only increase the computational complexity of networks, but also make no sense to the signs of binary features. To this end, Self-Distribution Binary Neural Network (SD-BNN) is proposed. Firstly, we utilize Activation Self Distribution (ASD) to adaptively adjust the sign distribution of activations, thereby improve the sign differences of the outputs of the convolution. Secondly, we adjust the sign distribution of weights through Weight Self Distribution (WSD) and then fine-tune the sign distribution of the outputs of the convolution. Extensive experiments on CIFAR-10 and ImageNet datasets with various network structures show that the proposed SD-BNN consistently outperforms the state-of-the-art (SOTA) BNNs (\eg, achieves 92.5\% on CIFAR-10 and 66.5\% on ImageNet with ResNet-18) with less computation cost. Code is available at \url{https://github.com/pingxue-hfut/SD-BNN}.
\end{abstract}

\section{Introduction}

\begin{figure*}
	\begin{center}
		\includegraphics[width=0.9\linewidth]{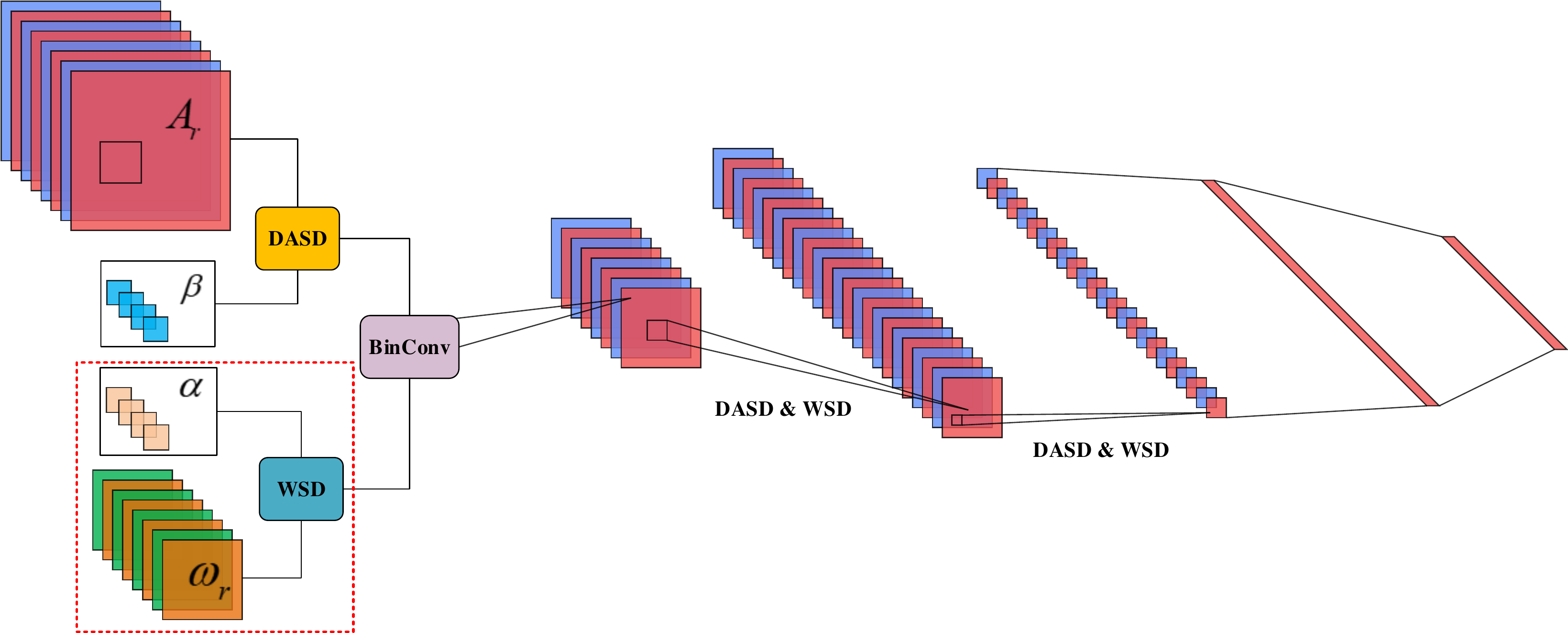}
	\end{center}
	\caption{Overview of SD-BNN (an example with LeNet \cite{31:726791}), consisting of Dynamic Activation Self Distribution (DASD) and Weight Self Distribution (WSD).}
	\label{fig 0}
\end{figure*}

With the development of deep learning, convolutional neural networks (CNNs) have been well demonstrated in a wide variety of computer vision applications, such as image classification \cite{1:nips/KrizhevskySH12,2:iccv/WangGYWY19}, object detection \cite{3:iccv/Girshick15,4:cvpr/PangCSFOL19}, and semantic segmentation \cite{5:iccv/HeD019,6:cvpr/ZhuangSTL019}. However, due to the massive parameters and high computational complexity in CNNs, it is difficult to deploy them in portable devices such as mobile phones and tablets. Since Courbariaux \etal \cite{7:nips/HubaraCSEB16} introduces binary neural networks (BNNs), network binarization is considered as one of the most promising solutions to this problem. On the one hand, compared with the conventional 32-bit floating-point neural networks, the BNNs represent network parameters with only 1-bit, which greatly reduces the storage requirement; on the other hand, both the weights and activations in BNNs are constrained to \{-1, +1\}, floating-point multiplication and addition operations with high computational complexity in convolution are replaced with low-cost XNOR and bitcount, which greatly accelerates the model inference process. Although network binarization achieves high model compression rate, it causes a significantly huge accuracy drop compared with full-precision models. Therefore, substantial research efforts are invested in minimizing the accuracy gap between BNNs and real-valued neural networks in recent years, and have made much progress. Among them, IR-NET \cite{8:cvpr/QinGLSWYS20} and BBG \cite{9:icassp/ShenLGH20} balance and normalize the weights to achieve maximum information entropy, reduce the information loss in the forward propagation process, and improve the network accuracy. BNN-DL \cite{10:cvpr/DingCLM19} regularizes the activation distribution for alleviating the problems of degeneration, saturation, and gradient mismatch caused by network binarization. CI-BCNN \cite{11:cvpr/WangLT0019} alleviates the inconsistency of signs in binary feature maps compared with their real-valued counterparts through channel-wise interactions. Inspired by the above methods, we argue that proper adjustment of sign distribution of weights and activations can improve the accuracy of BNNs without extra cost, and then Self-Distribution Binary Neural Network (SD-BNN) is proposed.\\
\indent
To improve the accuracy of BNNs, XNOR-Net \cite{12:eccv/RastegariORF16} introduces real-valued scaling factors into weights and activations, which effectively improves the performance of BNNs on large-scale datasets (\eg, ImageNet). After then, the scaling factors have been widely used in various optimization methods of network binarization \cite{13:bmvc/BulatT19,14:iclr/MartinezYBT20}. Binary weights and activations are multiplied by the scaling factors to approximate the real-valued weights and activations, through which the accuracy of BNNs is effectively improved. However, the floating-point multiplication operations caused by the scaling factors inevitably offset the speedup of network binarization partially. Besides, the scaling factors cannot adjust the sign distribution of weights and activations. To solve the above-mentioned problems, in SD-BNN (see the overview in Figure \ref{fig 0}), we replace the scaling factors with self-distribution factors, and then the multiplication operations associated with the scaling factors are replaced with addition operations, which not only obtains adaptive sign distribution adjustment, but also effectively reduces the computational complexity. Furthermore, the proposed methods are independent of network structures and training schemes, so they have good versatility and can be easily applied to almost all existing BNNs. Extensive experiments on CIFAR-10 \cite{15:pami/TorralbaFF08} and ImageNet \cite{16:cvpr/DengDSLL009} datasets with various network structures are conducted, and the experimental results show that the proposed SD-BNN consistently outperforms most state-of-the-art (SOTA) binarization methods.

\section{Related Work}

The network binarization has reached a very high model compression ratio. However, directly binarizing the models may cause large accuracy drops, especially on large-scale datasets. XNOR-Net \cite{12:eccv/RastegariORF16} achieves great performance improvement on the ImageNet dataset by introducing real-valued scaling factors. ABC-Net \cite{17:nips/LinZP17} suggests using a linear combination of multiple binary bases to approximate real-valued weights and activations to improve the accuracy. Bi-Real Net \cite{18:eccv/LiuWLYLC18} connects real-valued activations with binary activations to improve the representational capability of networks. HORQ-Net \cite{19:iccv/LiNZY017} reduces the residual between real-valued activations and binary activations by utilizing a high-order approximation scheme. CBCN \cite{20:cvpr/LiuDXZGLJD19} enhances the diversity of the intermediate feature maps by rotating the weight matrices, although no additional model parameters are added, it greatly increases the computational complexity of the networks. BinaryDuo \cite{21:iclr/KimK0K20} combines the advantages of the high compression rate of BNNs and the better accuracy of the ternary neural networks (TNNs) by decoupling TNNs into BNNs. BENN \cite{22:cvpr/ZhuDS19} uses multiple BNNs for ensembles to obtain higher accuracy. Although great progress has been made in network binarization, the existing BNNs still have a significant gap in accuracy compared with their full-precision counterparts. In addition, these methods have almost doubled the computational complexity or more besides the introduction of real-valued scaling factors, which largely offset the advantages of network binarization. To address these issues, we propose the SD-BNN, which eliminates the real-valued scaling factors and effectively improves the network performance without a large amount of additional computation. At the same time, our methods can be conveniently applied to almost all existing BNNs to further reduce the accuracy gap between BNNs and real-valued networks.

\section{Method}

In this section, we first review the preliminaries of BNNs (see section \ref{3.1}). Then, we present the details of SD-BNN and its training, including Activation Self Distribution (see section \ref{3.2}), improved Activation Self Distribution (see section \ref{3.3}), Weight Self Distribution (see section \ref{3.4}), and the optimization methods of SD-BNN with self-distribution factors (see section \ref{3.5}).	


\subsection{Binary Neural Networks}
\label{3.1}

Convolution is one of the most important operations in CNNs, which can be formalized as: 
\begin{equation}
	z = {\omega _r} \otimes {A_r}
\end{equation}
where ${\omega _r}$ indicates the real-valued weights, ${A_r}$ indicates the real-valued input activations, and $ \otimes $ indicates the convolution. BNNs represent the network weights and activations with 1-bit, so it is necessary to binarize the real-valued weights and activations, and we usually use the sign function to get binarization, which is presented as: 
\begin{equation}
	\begin{split}
		sign(x) &= \left\{ \begin{array}{l}
			+ 1,{\kern 1pt} \;\;if\;x \ge 0\\
			- 1,\;\;otherwise
		\end{array} \right.\\
		{\omega _b} &= sign({\omega _r})\\
		{A_b} &= sign({A_r})
	\end{split}
\end{equation}
where ${\omega _b}$ and ${A_b}$ denote the binary weights and activations, respectively. While the real-valued weights and activations are binarized, most BNNs introduce real-valued scaling factors to improve the network accuracy, so the convolution in BNNs is defined as:
\begin{equation}
	z = {\alpha _s}{\beta _s}({\omega _b} \oplus {A_b})
\end{equation}
where ${\alpha _s}$ and ${\beta _s}$ denote the scaling factors of weights and activations respectively. $ \oplus $ denotes the inner production for tensors with bitwise operations XNOR and bitcount. ${\alpha _s}$ and ${\beta _s}$ can be calculated analytically \cite{12:eccv/RastegariORF16}, or be learned discriminatively via  backpropagation \cite{13:bmvc/BulatT19}.


\subsection{Activation Self Distribution}
\label{3.2}

\begin{figure}[t]
	\centering
	\subfigure[Original $\beta $]{
		\includegraphics[width=0.38\linewidth]{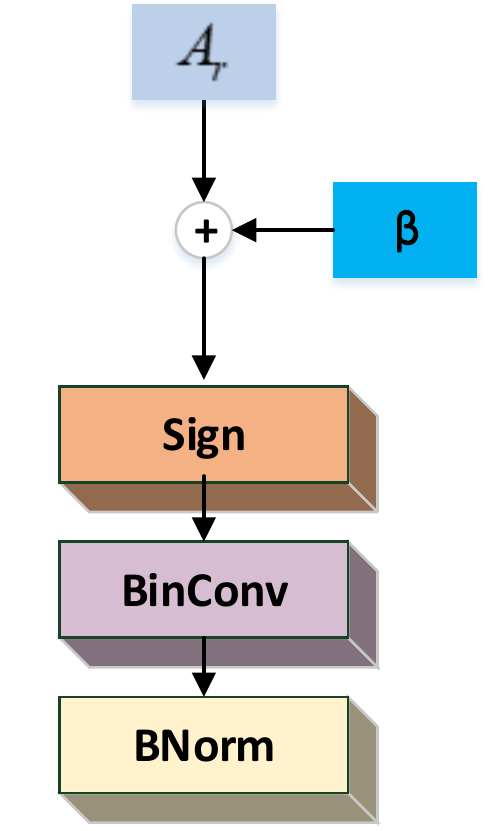}
	}
	\subfigure[$\beta {\rm{ = }}{\mathop{\rm Tanh}\nolimits} (\beta )$]{
		\includegraphics[width=0.53\linewidth]{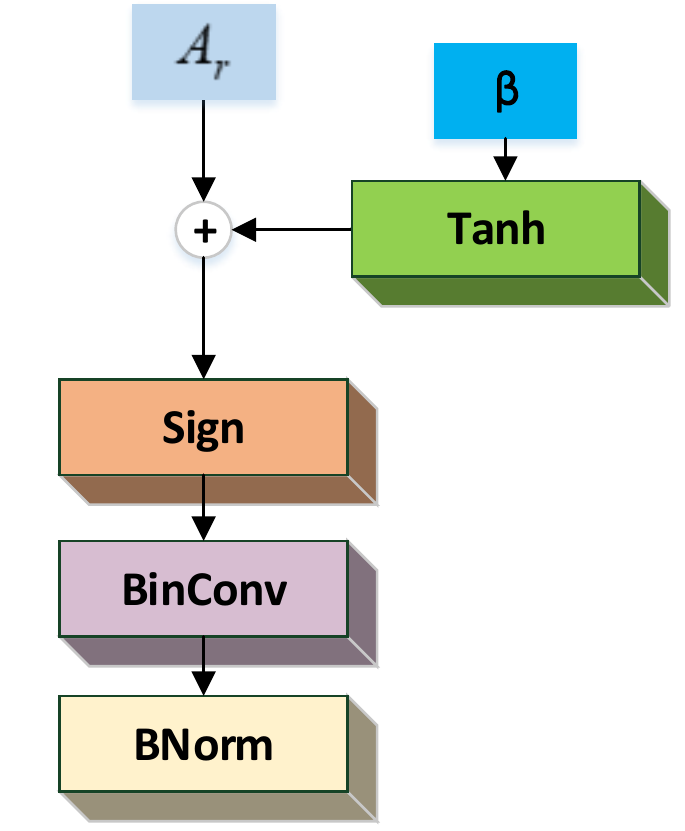}
	}
	\subfigure[$\beta {\rm{ = Sigmoid}}(\beta )$]{
		\includegraphics[width=0.53\linewidth]{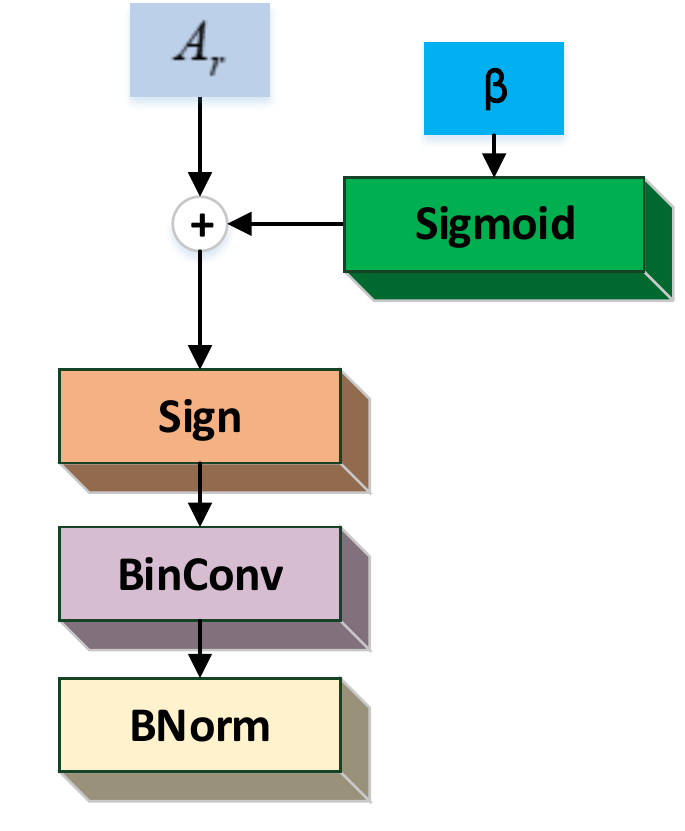}
	}
	\caption{Three forms of the ASD factor $\beta $.}
	\label{fig 1}
\end{figure}

\begin{figure}[t]
	\begin{center}
		\includegraphics[width=1\linewidth]{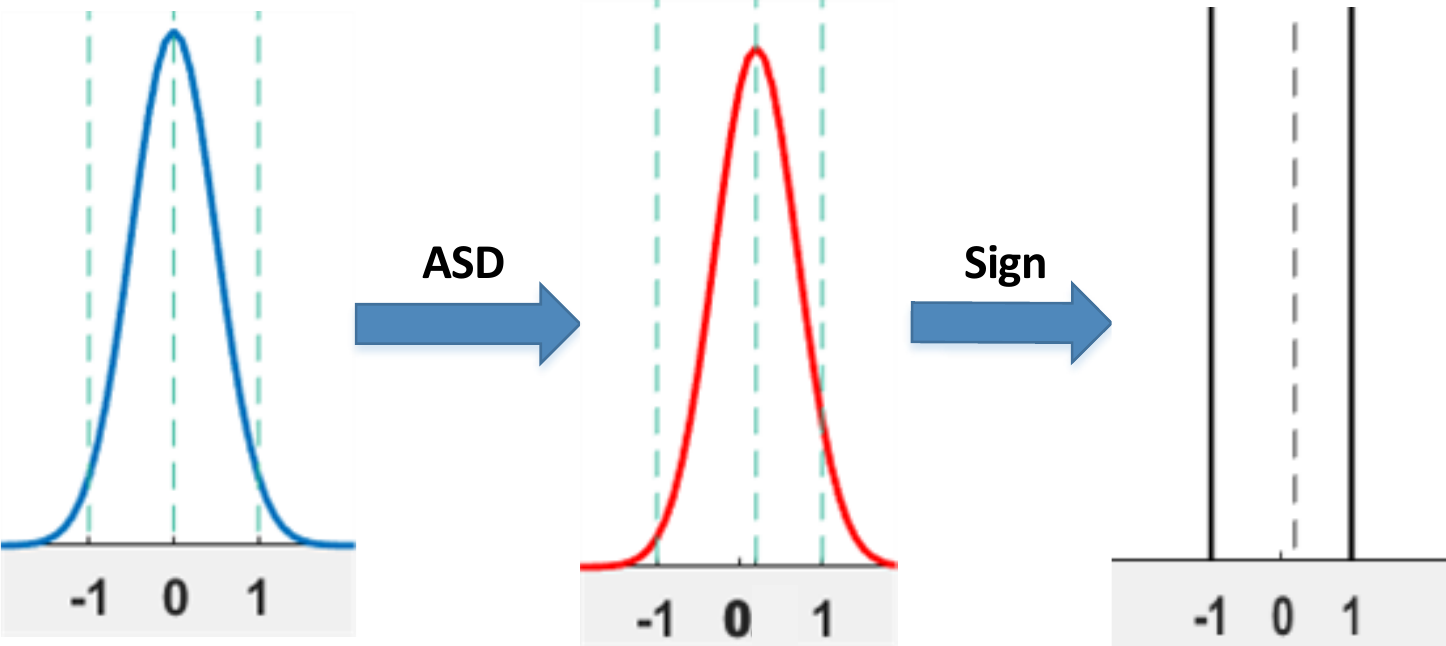}
	\end{center}
	\caption{An illustration of the effect of ASD on the sign distribution of activations.}
	\label{fig 2}
\end{figure}

Different from prior work, we remove the scaling factors (which can also be retained for being compatible with existing BNNs) of weights and activations, and redefine $\alpha $ and $\beta $ as self-distribution factors, where $\alpha $ is the Weight Self Distribution (WSD) factor, and $\beta $ the Activation Self Distribution (ASD) factor. This section mainly presents the ASD method, and WSD will be described in section \ref{3.4}.\\
\indent
BNN-DL \cite{10:cvpr/DingCLM19} defines three problems: (1) degeneration: for the input activations ${A_r}$ and the value $v \in {A_r}$, the signs of all the values $sign(v)$ are the same; (2) saturation: almost all $v \in {A_r}$, $|v| > 1$; or (3) gradient mismatch: almost all $v \in {A_r}$, $|v| < 1$. They argue that the difficulty of training BNNs is mainly caused by the above problems, and propose to regularize the activation distribution by adding regularization terms to the loss function to alleviate those problems, thereby improve the training stability and the network accuracy. CI-BCNN \cite{11:cvpr/WangLT0019} finds that binary convolution usually obtains inconsistent signs in binary feature maps compared with their full-precision counterparts, \ie, $sign({\omega _r} \otimes {A_r}) \ne sign(sign({\omega _r}) \oplus sign({A_r}))$, and this inconsistency leads to significant information loss. Therefore, they guide the binary feature maps to learn the sign distribution of the real-valued feature maps through the teacher-student interactions. We apply a simple yet effective way to directly adjust the sign distribution of weights and activations.\\
\indent
Given a L-layer CNN model, let ${A_r} \in {R^{C \times H \times W}}$ be the input activations of the ${l_{th}}$ layer, where $C$, $H$ and $W$ represents the input channels, height and width respectively. Considering the trade-off between accuracy and efficiency, following the settings of the scaling factors, the self-distribution factors adjust the sign distribution of activations/weights also in the channel-wise way. Then, we define the ASD factor $\beta  \in {R^C}$, which is learned via backpropagation, and the ASD can be expressed as:
\begin{equation}
	\begin{split}
		\;\;{A_r} &= {A_r} + \beta \\
		{A_b} &= sign({A_r})\\
		\;\;z &= {\omega _b} \oplus {A_b}
	\end{split}
\end{equation}
\indent
It is worth noting that, in contrast to the scaling factors acting on the weights/activations after binarization, the self-distribution factors work before binarization, so that which can effectively adjust the sign distribution of activations/weights while avoiding real values from participating in the convolution.\\
\indent
Given that in BNNs, after one of the various activation functions, the values of the input activations $v \in {A_r}$ are mainly distributed as [-1,1] or [0,1], \ie, almost all $\left| v \right| \in [0,1]$. An intuitive assumption that if $\left| \beta  \right| > 1$, most signs of the values $sign(v)$ could be the same after the self-distribution operation, which would severely break the balance and diversity of the activation distribution, and eventually affect the performance of the networks. To this end, we have considered the following three forms of the ASD factor:\\
\indent
(a) Original $\beta $, \ie, there is no constraint on the value range and the sign of $\beta $.\\
\indent
(b) $\beta {\rm{ = }}{\mathop{\rm Tanh}\nolimits} (\beta )$, then $\beta  \in [ - 1,1]$, there is no constraint on the sign of $\beta $.\\
\indent
(c) $\beta {\rm{ = Sigmoid}}(\beta )$, then $\beta  \in [0,1]$, both the value range and the sign of $\beta $ are constrained.\\
\indent
Figure \ref{fig 1} shows the three forms of $\beta $ and experiments have proved the above assumption that when no constraint on the value range and/or sign of $\beta $, the loss fluctuates during the training and the network is difficult to converge, which finally affects the network accuracy (see section \ref{4.2}). Therefore, in practical applications, we recommend using the form shown in Figure \ref{fig 1}(c).\\
\indent
Figure \ref{fig 2} shows the effect of ASD on the activation distribution. We use the learnable parameters $\beta $ and the activations ${A_r}$ for addition operation to shift the activation distribution, and then the sign distribution of activations changes. The activations after binarization participate in the convolution, which affect the sign distribution of the outputs.


\subsection{Dynamic Activation Self Distribution}
\label{3.3}

\begin{figure}[t]
	\begin{center}
		\includegraphics[width=0.78\linewidth]{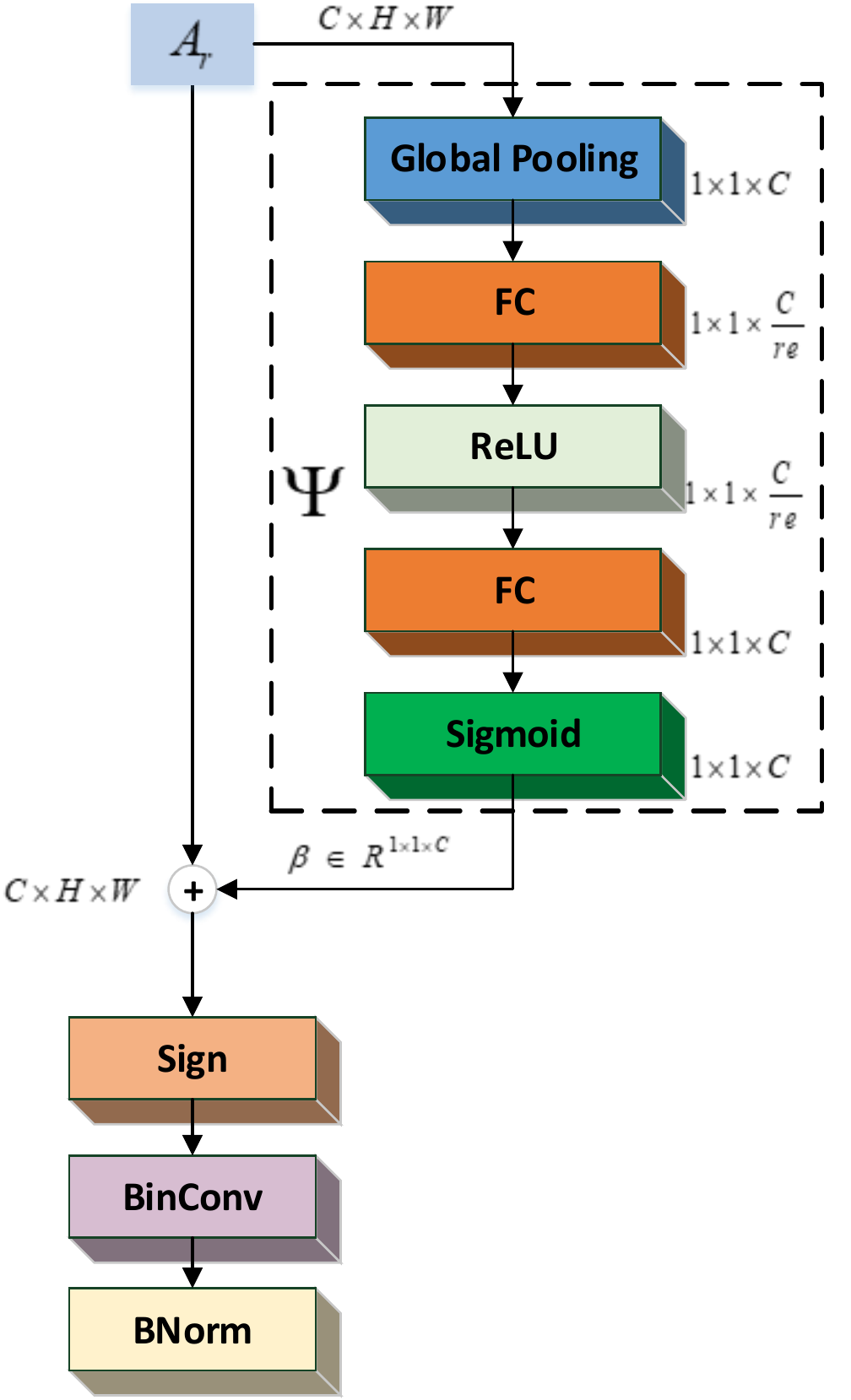}
	\end{center}
	\caption{The schematic of DASD.}
	\label{fig 3}
\end{figure}

In a neural network, the activations are dynamic and related to the inputs. Although ASD can effectively adjust the sign distribution of activations, the self-distribution factor $\beta $ would be constant and static when the network training is done (called static ASD), so the ASD may have limitations. Therefore, we argue that if $\beta $ could be dynamic just the same as activations which are changed with different inputs, then the performance of BNNs could be further improved. So Dynamic Activation Self Distribution (DASD) is designed, which is defined as:
\begin{equation}
	\begin{array}{l}
		\beta {\rm{ = }}\Psi ({A_r},{\omega _\Psi },re)\\
		\;\;{A_r} = {A_r} + \beta \\
		{A_b} = sign({A_r})
	\end{array}
\end{equation}
where ${\omega _\Psi }$ denotes the parameters of the function $\Psi $ which takes the activations as input. And $re$ is a hyperparameter, which controls the number of parameters in $\Psi $. Function $\Psi $ maps the input activations ${A_r}$ to the self-distribution factor $\beta $, and now $\beta $ is dynamic. 
Figure \ref{fig 3} shows our implementation of function $\Psi $. The design is inspired by \cite{23:pami/HuSASW20}, but we use $\Psi $ to calculate the ASD factor dynamically rather than as a gating function for obtaining scaling factors of activations \cite{14:iclr/MartinezYBT20} or as a self-attention mechanism \cite{23:pami/HuSASW20}.


\subsection{Weight Self Distribution}
\label{3.4}

\begin{figure}[t]
	\begin{center}
		\includegraphics[width=0.7\linewidth]{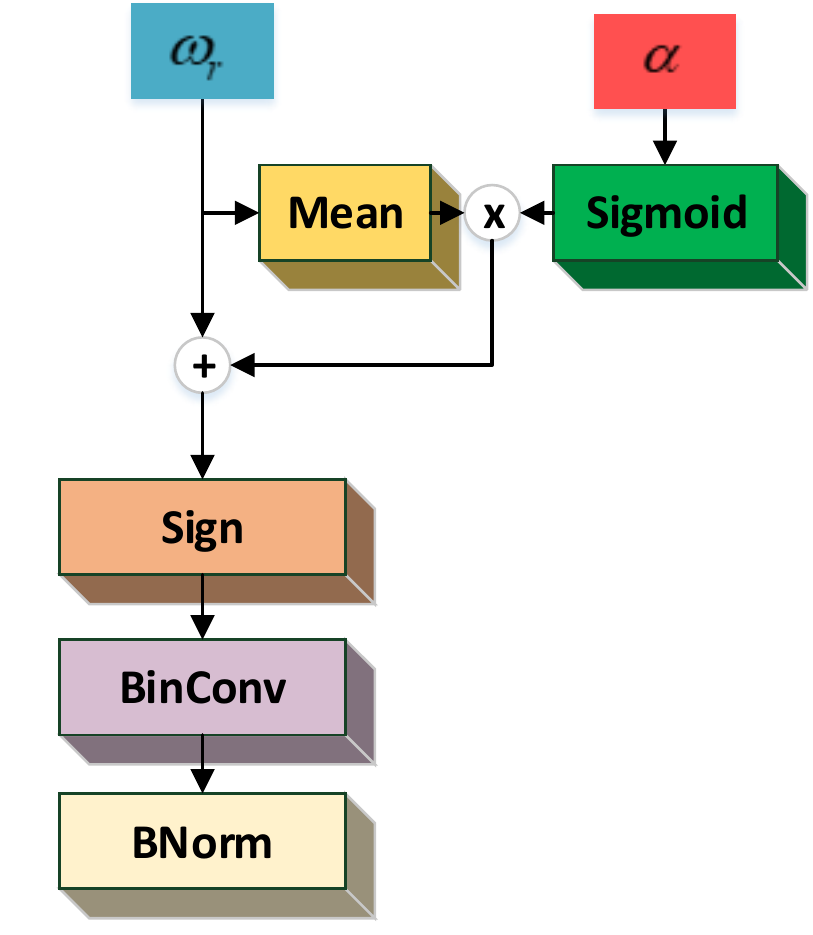}
	\end{center}
	\caption{The schematic of WSD.}
	\label{fig 4}
\end{figure}

\begin{figure*}[t]
	\begin{center}
		\includegraphics[width=0.6\linewidth]{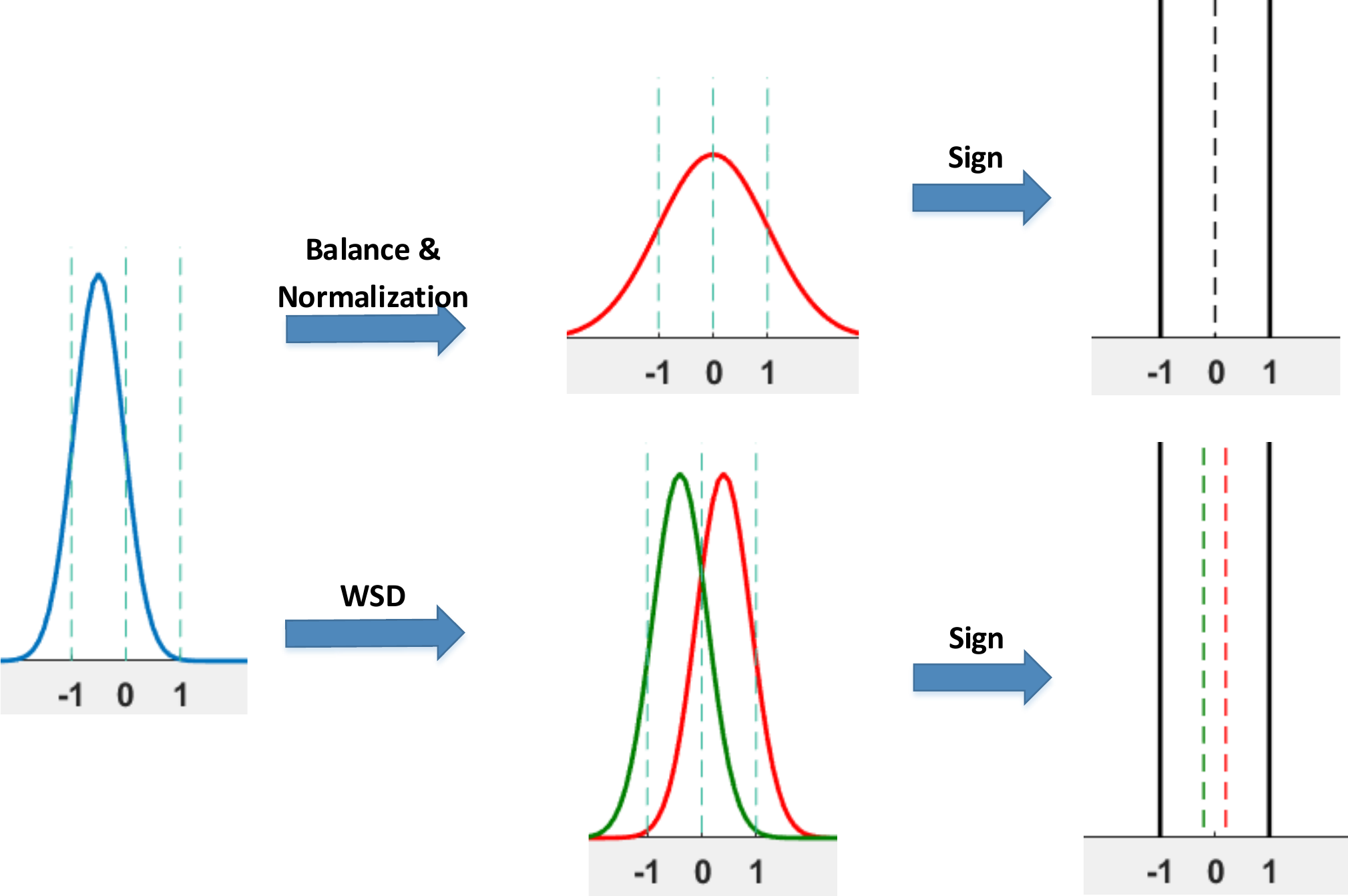}
	\end{center}
	\caption{The differences in the effect of WSD and “Balance \& Normalization” on the sign distribution of weights.}
	\label{fig 5}
\end{figure*}

IR-Net \cite{8:cvpr/QinGLSWYS20} and BBG \cite{9:icassp/ShenLGH20} argue that balancing and normalizing the weights (which makes the mean of weights as 0 and the standard deviation as 1) could maximize the information entropy and effectively reduces the information loss in the forward propagation process, and thereby improves the network accuracy. In this work, we review the weight distribution from the perspective of their signs instead. Although the magnitude of the real-valued weights is changed after normalization, the signs of the weights are not changed. After weight binarization, the magnitude of the real-valued weights becomes meaningless for the convolution. Balancing the weights would change the sign distribution and the number of positive and negative values in the weights would tend to be the same (but not absolutely the same because the magnitude of each value in the real-valued weights is not equal compared with the binary weights). We argue that such sign balance may be suboptimal, so we propose WSD.\\
\indent
In the ASD method, the sign distribution could be changed through using the ASD factor $\beta $ on the input activations ${A_r}$. The modified ${A_r}$ would participate in the convolution after binarization, thereby indirectly affects the sign distribution of the outputs. Similarly, we introduce the learnable WSD factor $\alpha $ in WSD method. We propose to change the sign distribution of weights ${\omega _r} \in {R^{Cout \times Cin \times K \times K}}$ by using $\alpha  \in {R^{Cout}}$ (where $Cout$ and $Cin$ represent the number of output and input channels, $K$ the width and height of the kernel), and then to affect the sign distribution of the outputs through the convolution. Besides, similar to the ASD factor $\beta $, the value range of $\alpha $ also needs to be under consideration. In order to ensure $\alpha $ a comparable magnitude with the weights, we use the mean of weights as a benchmark, so WSD is formulated as:
\begin{equation}
	\begin{array}{l}
		{\overline \omega  _r} = Mean({\omega _r}) \in {R^{Cout}}\\
		\alpha  = sigmoid(\alpha ) \times \overline \omega _r \\
		\;\;\;\;\;\;{\omega _r} = {\omega _r} + \alpha \\
		\;\;\;\;{\omega _b} = sign({\omega _r})
	\end{array}
\end{equation}
where $Mean({\omega _r})$ is used to calculate the mean of weights ${\omega _r}$ channel-wise. Figure \ref{fig 4} describes the implementation of WSD, and Figure \ref{fig 5} shows the differences in the effect of WSD and “Balance \& Normalization” on the sign distribution of weights.

\indent
Compared with the ASD method, WSD has a greater advantage in the network inference speedup. During the inference stage, the ASD factor $\beta $ is still stored in the network model and participates in forward propagation. However, once the training is completed, the weights do not change, and the model only retains the weights after binarization. Therefore, the calculation associated with the WSD factor $\alpha $ would not reduce the model inference speed.


\subsection{Training}
\label{3.5}

Our SD-BNN only adjusts the sign distribution of weights and activations, and is independent of network structures and training methods, so the existing training methods of BNNs could be directly applied. In the backward propagation, the derivative of the sign function is zero almost everywhere, so Straight-Through Estimator (STE) \cite{24:corr/BengioLC13} and its variants \cite{25:iccv/GongLJLHLYY19,26:aaai/HuangNY19,18:eccv/LiuWLYLC18} could be used. And then, the self-distribution factors introduced in this work would be optimized together with the network weights via backpropagation. Specifically, $\alpha $ and $\beta $ are updated as follows:
\begin{equation}
	{\alpha ^{t + 1}} = {\alpha ^t} - \eta \frac{{\partial L}}{{\partial {\alpha ^t}}} = {\alpha ^t} - \eta \frac{{\partial L}}{{\partial {\omega _b}}}\frac{{\partial {\omega _b}}}{{\partial {\omega _r}}}\frac{{\partial {\omega _r}}}{{\partial {\alpha ^t}}}
\end{equation}

\begin{equation}
	{\beta ^{t + 1}} = {\beta _t} - \eta \frac{{\partial L}}{{\partial {\beta ^t}}} = {\beta ^t} - \eta \frac{{\partial L}}{{\partial {A_b}}}\frac{{\partial {A_b}}}{{\partial {A_r}}}\frac{{\partial {A_r}}}{{\partial {\beta ^t}}}
\end{equation}
where $t$ denotes the number of current iteration, $\eta $ represents the learning rate and $L$ the loss, $\frac{{\partial {\omega _b}}}{{\partial {\omega _r}}}$ and $\frac{{\partial {A_b}}}{{\partial {A_r}}}$ are calculated approximately using STE or its variants respectively. The training process of our SD-BNN is summarized in Algorithm \ref{al 1}.

\begin{algorithm}[!ht]
	\caption{Forward and backward propagation for the training of the proposed SD-BNN.}
	\label{al 1}
	\textbf{Require}: the real-valued input activations ${A_r}$, the real-valued weights ${\omega _r}$, the ASD factor $\beta $ and the WSD factor $\alpha $. \\
	\textbf{Forward propagation} \\
	\quad Adjust the sign distribution of ${A_r}$ and then binarize: 
	\[\begin{array}{l}
		{A_r} = {A_r} + \beta \\
		{A_b} = sign({A_r})
	\end{array}\] \\
	\quad Adjust the sign distribution of ${\omega _r}$ and then binarize: 
	\[\begin{array}{l}
		{\omega _r} = {\omega _r} + \alpha \\
		{\omega _b} = sign({\omega _r})
	\end{array}\] \\
	\quad Calculate th outputs:  $z = {\omega _b} \oplus {A_b}$ \\
	\textbf{Back propagation} \\
	\quad Calculate the gradients w.r.t. ${A_r}$:
	$\frac{{\partial L}}{{\partial {A_r}}} = \frac{{\partial L}}{{\partial {A_b}}}\frac{{\partial {A_b}}}{{\partial {A_r}}}$ \\
	\quad Calculate the gradients w.r.t. ${\omega _r}$:
	$\frac{{\partial L}}{{\partial {\omega _r}}} = \frac{{\partial L}}{{\partial {\omega _b}}}\frac{{\partial {\omega _b}}}{{\partial {\omega _r}}}$ \\
	\quad Calculate the gradients w.r.t. $\beta $:
	$\frac{{\partial L}}{{\partial \beta }}{\rm{ = }}\frac{{\partial L}}{{\partial {A_r}}}\frac{{\partial {A_r}}}{{\partial \beta }}$ \\
	\quad Calculate the gradients w.r.t. $\alpha $:
	$\frac{{\partial L}}{{\partial \alpha }}{\rm{ = }}\frac{{\partial L}}{{\partial {\omega _r}}}\frac{{\partial {\omega _r}}}{{\partial \alpha }}$ \\
	\textbf{Parameters Update} \\
	\quad Update ${\omega _r} :{\omega _r}  = {\omega _r}  - \eta \frac{{\partial L}}{{\partial {\omega _r}}}$ \\
	\quad Update $\beta :\beta  = \beta  - \eta \frac{{\partial L}}{{\partial \beta }}$ \\
	\quad Update $\alpha :\alpha  = \alpha  - \eta \frac{{\partial L}}{{\partial \alpha }}$
\end{algorithm}


\section{Experiments}
To verify the effectiveness of the proposed SD-BNN, we conduct experiments on two benchmark datasets CIFAR-10 and ImageNet (ILSVRC12) with various network structures such as VGG-Small \cite{27:eccv/ZhangYYH18}, ResNet-20 and ResNet-18 \cite{28:cvpr/HeZRS16}.\\
\indent
In this section, we first describe the implementation details of the experiments (see section \ref{4.1}). Then, to evaluate the effect of WSD, ASD, and DASD proposed in this work, we conduct ablation analysis (see section \ref{4.2}). Finally, we implement comprehensive experiments to compare our SD-BNN with other SOTA methods to verify the superiority of the proposed methods (see section \ref{4.3}, \ref{4.4}).  


\subsection{Implementation Details}
\label{4.1}

Since the proposed SD-BNN does not rely on any specified training methods of BNNs, existing training methods are all applicable theoretically. In the experiments, we implement SD-BNN based on PyTorch. In terms of binarization, it is consistent with other BNNs that we binarize all other convolutional layers except for the first and last layers of the networks. In terms of data augmentation, we use the same operations as the existing BNNs do on the datasets. In terms of network training, to train the networks from scratch, we use the EDE method during the backpropagation process which is proposed in IR-Net \cite{8:cvpr/QinGLSWYS20} as our baseline on CIFAR-10, without using any pre-trained models; and for ImageNet, we use the training methods proposed by \cite{29:corr/abs-1904-05868}. For the experiments using the DASD method, the default setting of the hyperparameter $re$ in $\Psi $ as $re = 16$. We mostly follow the settings of their original papers \cite{18:eccv/LiuWLYLC18,8:cvpr/QinGLSWYS20,12:eccv/RastegariORF16,27:eccv/ZhangYYH18}, including initialization, the order of the blocks, and hyperparameter selection, etc. if without otherwise specified.


\subsection{Ablation Study}
\label{4.2}

To investigate the performance of WSD and ASD respectively, and to evaluate the consistency of the superposition effect of the two methods and verify the assumption on the constraint on the ASD factor $\beta $, we perform ablation study on the CIFAR-10 dataset across VGG-Small, ResNet-20, and ResNet-18, and the results are shown in Table \ref{tab 1}.

\begin{table*}
	\begin{center}
		\begin{tabular}{|c|c|c|c|c|}
			\hline
			\multirow{2}*{Method} &
			\multirow{2}*{Bit-width(W/A)} &
			\multicolumn{3}{c|}{Accuracy(\%)}\\
			\cline{3-5}
			& &VGG-Small&ReNet-20&ReNet-18 \\
			\hline\hline
			Full precision & 32/32 & 91.7 & 90.8 & 93.0 \\
			Baseline & 1/1 & 88.7 & 85.2 & 90.0 \\
			ASD (with sigmoid) & 1/1 & - & 86.2 & - \\
			ASD (with tanh) & 1/1 & - & 85.9 & - \\
			ASD (original) & 1/1 & - & 85.6 & - \\
			DASD & 1/1 & 90.7 & 86.9 & 92.3 \\
			WSD & 1/1 & 90.6 & 85.7 & 90.6 \\
			SD-BNN(W\&DASD) & 1/1 & $\bm{90.8}$ & $\bm{86.9}$ & $\bm{92.5}$ \\
			\hline
		\end{tabular}
	\end{center}
	\caption{Ablation study for SD-BNN.}
	\label{tab 1}
\end{table*}

\indent
\textbf{ASD}: We apply the ASD method to ResNet-20. From Table \ref{tab 1}, we can see that there is a large gap in whether using a constraint on the ASD factor. Compared with the baseline, it obtains a 1\% absolute accuracy increase when $\beta $ is constrained, while only a 0.4\% increase if not. Furthermore, the DASD has a better generalization ability than the static ASD does, which is 1.7\% higher than the baseline in terms of absolute accuracy. Besides, we also apply the DASD method to VGG-Small and ResNet-18. The results show that DASD can consistently obtain improvement with various network structures.\\
\indent
\textbf{WSD}: Similarly, we have applied the WSD method to each of the three network structures. Although the performance of WSD is not as good as DASD in terms of accuracy, its advantage is that once the model training is completed, the WSD method would not bring any extra storage and computation cost. Besides, we speculate that the performance differences between WSD and DASD is caused by the following reasons:\\
\indent
(1) When the model training is completed, the network weights would be constant, and the WSD factor $\alpha $ does not participate in any calculations and could be removed safely. Therefore, similar to ASD, WSD is static to the inputs, while DASD is dynamic.\\
\indent
(2) The signs of weights and input activations have different effect on the outputs of the binary convolution. If the sign of an input value is changed by ASD/DASD, it would affect the whole outputs. While if the sign of a certain weight value is changed by WSD, it only affects the outputs of the related channel. For intuitive understanding, Table \ref{tab 2} shows our reasoning process. This also explains that the performance of ASD is still better than that of WSD while both methods are static shown in Table \ref{tab 1}.

\begin{table*}
	\begin{center}
		\begin{tabular}{l}
			\hline
			The weights ${\omega _r} \in {R^{Cout \times Cin \times K \times K}}$ and the input activations ${A_r} \in {R^{Cin \times H \times W}}$. \\
			Generally $Cout \ge Cin$, for simplicity, assuming that $Cout = 2$, $Cin, K, H, W = 1$: \\
			\hline
			Without any methods: \\
			\qquad $sign({\omega _r}){\rm{ = [1,1]}}$, $sign({A_r}) = [1]$ \\
			\qquad $Output = sign({\omega _r}) \oplus sign({A_r}) = {\rm{[1,1]}}$ \\
			\hline
			Using DASD/ASD: \\
			\qquad $sign({\omega _r}){\rm{ = [1,1]}}$, $sign({A_r} + \beta ) = [ - 1]$ \\
			\qquad  $Output = sign({\omega _r}) \oplus sign({A_r} + \beta ) = {\rm{[ - 1, - 1]}}$ \\
			\hline
			Using WSD: \\
			\qquad $sign({\omega _r} + \alpha ){\rm{ = [ - 1,1]}}$, $sign({A_r}) = [1]$ \\
			\qquad $Output = sign({\omega _r} + \alpha ) \oplus sign({A_r}) = {\rm{[ - 1,1]}}$ \\
			\hline
		\end{tabular}
	\end{center}
	\caption{Analysis of the differences between WSD and ASD/DASD.}
	\label{tab 2}
\end{table*}

\textbf{SD-BNN}: Although the performance of WSD is not as good as ASD/DASD, the adjustment granularity of WSD is finer, so the networks could be further tuned through WSD after using DASD. Table \ref{tab 1} shows that the improvement of the two methods could be superimposed and have consistency across various network structures.

\subsection{Comparison with SOTA Methods}
\label{4.3}

In this section, based on the image classification tasks, we further evaluate the proposed SD-BNN on CIFAR-10 and ImageNet datasets with various network structures by comparing it with existing SOTA methods, including BNN \cite{7:nips/HubaraCSEB16}, XNOR-Net \cite{12:eccv/RastegariORF16}, DSQ \cite{25:iccv/GongLJLHLYY19}, BNN-DL \cite{10:cvpr/DingCLM19}, IR-Net \cite{8:cvpr/QinGLSWYS20}, BinaryDuo \cite{21:iclr/KimK0K20}, CI-BCNN \cite{11:cvpr/WangLT0019}, BBG \cite{9:icassp/ShenLGH20}, ABC-Net \cite{17:nips/LinZP17}, Bi-Real Net \cite{18:eccv/LiuWLYLC18}, XNOR-Net++ \cite{13:bmvc/BulatT19}, and Real-to-Bin \cite{14:iclr/MartinezYBT20}.\\
\indent
\textbf{Comparison on CIFAR-10}: The CIFAR-10 dataset consists of 60,000 images of size 32x32, which are divided into 10 categories, of which 50,000 are the training set and the rest 10,000 are the test set. Table \ref{tab 3} lists the performance using different methods on CIFAR-10, and shows that in all cases, the proposed SD-BNN obtains the best accuracy. Moreover, our SD-BNN with the original ResNet structure \cite{28:cvpr/HeZRS16} even surpasses the IR-Net which uses the Bi-Real Net structure \cite{18:eccv/LiuWLYLC18}. Over ResNet-18, we narrow the absolute accuracy gap between the binary one and its full-precision counterpart to 0.5\%.

\begin{table}
	\begin{center}
		\scalebox{0.92}{
			\begin{tabular}{|c|c|c|c|}
				\hline
				Model & Method & Bit-width(W/A) & Accuracy(\%) \\
				\hline\hline
				\multirow{7}*{VGG-Small} & Full-precision & 32/32 & 91.7 \\
				&XNOR-Net & 1/1 & 89.8 \\
				&BNN & 1/1 & 89.9 \\
				&BNN-DL & 1/1 & 90.0 \\
				&IR-Net & 1/1 & 90.4 \\
				&BinaryDuo & 1/1 & 90.4 \\
				&Ours & 1/1 & $\bm{90.8}$ \\ 
				\hline
				\multirow{5}*{ResNet-20} & Full-precision & 32/32 & 90.8 \\
				&DSQ & 1/1 & 84.1 \\
				&IR-Net & 1/1 & 85.4 \\
				&IR-Net* & 1/1 & 86.5 \\
				&Ours & 1/1 & $\bm{86.9}$ \\ 
				\hline
				\multirow{4}*{ResNet-18} & Full-precision & 32/32 & 93.0 \\
				&BNN-DL & 1/1 & 90.5 \\
				&IR-Net & 1/1 & 91.5 \\
				&Ours & 1/1 & $\bm{92.5}$ \\ 
				\hline
		\end{tabular}}
	\end{center}
	\caption{Accuracy comparison with state-of-the-art methods on CIFAR-10 (*These methods use Bi-Real Net structure).}
	\label{tab 3}
\end{table}

\textbf{Comparison on ImageNet}: Compared with CIFAR-10, ImageNet (ILSVRC12) is more challenging because of its larger scale and more diverse categories. ImageNet contains approximately 1.2 million training images and 50,000 validation images from 1,000 categories. Therefore, we study the performance of SD-BNN on ImageNet based on ResNet-18 to further verify its effectiveness on large-scale classification tasks. Similar to Real-to-Bin \cite{14:iclr/MartinezYBT20}, we use the settings and training methods proposed by \cite{29:corr/abs-1904-05868}, including using PReLU \cite{30:iccv/HeZRS15} as activation function, reverse-order initialization, and knowledge distillation. Table \ref{tab 4} shows the performance comparison between the existing SOTA methods and our SD-BNN. It can be seen from the table that our SD-BNN outperforms all other methods by large margins. After removing the scaling factors of weights and activations, SD-BNN still obtains a 1.1\% increase in terms of the absolute accuracy on Top-1 compared with Real-to-Bin \cite{14:iclr/MartinezYBT20}. The experimental results show that our SD-BNN consistently outperforms other SOTA BNNs on the datasets with various scales.

\begin{table}
	\begin{center}
		\scalebox{0.98}{
			\begin{tabular}{|c|c|c|c|}
				\hline
				Method & Bit-width(W/A) & Top-1(\%) & Top-5(\%) \\
				\hline\hline
				Full-precision & 32/32 & 69.6 & 89.2 \\
				ABC-Net & 1/1 & 42.7 & 67.6 \\
				XNOR-Net & 1/1 & 51.2 & 73.2 \\
				Bi-real Net & 1/1 & 56.4 & 79.5 \\
				XNOR-Net++ & 1/1 & 57.1 & 79.9 \\
				IR-Net* & 1/1 & 58.1 & 80.0 \\
				BGG* & 1/1 & 59.4 & - \\
				CI-BCNN* & 1/1 & 59.9 & 84.2 \\
				BinaryDuo* & 1/1 & 60.9 & 82.6 \\
				Real-to-Bin*$\dag $ & 1/1 & 65.4 & 86.2 \\
				Ours*$\dag $ & 1/1 & $\bm{66.5}$ & $\bm{86.7}$ \\ 
				Ours*$\dag $ & 32/1 & 67.6 & 87.4 \\
				\hline
		\end{tabular}}
	\end{center}
	\caption{Accuracy comparison with state-of-the-art methods on ImageNet (*These methods use Bi-Real Net structure, $\dag re = 8$).} 
	\label{tab 4}
\end{table}


\subsection{Complexity Analysis}
\label{4.4}

We further compare the computation cost of the existing SOTA methods and our SD-BNN on different datasets respectively. We use 1 Nvidia RTX 2070 GPU for model training and validation on CIFAR-10, and 2 Nvidia RTX Titan GPUs while on ImageNet. Table \ref{tab 5} shows the average delay per batch size in the validation process of our methods and other SOTA methods. The results show that replacing the widely used scaling factors with the self-distribution factors proposed in this work can effectively reduce the computational complexity of the model, thereby speed up model inference.

\begin{table}
	\begin{center}
		\scalebox{0.8}{
			\begin{tabular}{|c|c|c|c|}
				\hline
				Dataset  & Method & Bit-width(W/A) & Time/batch size(ms) \\
				\hline\hline
				CIFAR-10 & IR-Net & 1/1 & 73 \\
				batch size=128 & Ours & 1/1 & $\bm{68}$ \\
				\hline
				ImageNet & Real-to-Bin*$\dag $ & 1/1 & 572 \\
				batch size=512 &Ours*$\dag $ & 1/1 & $\bm{540}$ \\
				\hline
			\end{tabular}
		}
	\end{center}
	\caption{Comparison of time cost of ResNet-18 with different methods (*These methods use Bi-Real Net structure, $\dag re = 8$).} 
	\label{tab 5}
\end{table}


\section{Conclusion}

Given that in BNNs, the outputs of the convolution are only affected by the signs of weights and input activations. In this work, we review the scaling factors of weights and activations which are widely used in BNNs, replace them with self-distribution factors, and propose ASD and WSD which could directly and effectively change the sign distribution of activations and weights. Extensive experiments have proved that our SD-BNN has consistent superiority compared with the existing SOTA methods on different datasets and with various network structures. Besides, since the scaling factors are replaced with the self-distribution factors, during the inference process, SD-BNN replaces the floating-point multiplication operations of activations and scaling factors existing in other BNNs with addition operations, and eliminates the computation cost brought by the multiplication operations of weights and scaling factors, and effectively reduces the model inference delay.


\section*{Acknowledgments}

This work was supported in part by the National Key Research and Development Program under Grant 2018YFC0604404, in part by the National Natural Science Foundation of China under Grant 61806067, and in part by the Anhui Provincial Key R\&D Program (202004a05020040).

{\small
\bibliographystyle{ieee_fullname}
\bibliography{egbib}
}

\end{document}